\documentclass[letterpaper, 10pt, conference]{ieeeconf}      %

\IEEEoverridecommandlockouts                              %

\usepackage[pdftex]{graphicx}
\graphicspath{{figures/}}
\usepackage{hyperref}
\hypersetup{
    colorlinks=true,
    linkcolor=black,
    citecolor=black,
    filecolor=black,
    urlcolor=black,
}
\usepackage[T1]{fontenc}  
\usepackage[utf8]{inputenc}
\usepackage[english]{babel}
\usepackage[export]{adjustbox}
\usepackage{caption}
\usepackage{subcaption}
\usepackage[colorinlistoftodos]{todonotes}
\usepackage{lipsum}

\usepackage{placeins}%

\title{\LARGE \bf
Cooperative Motion Planning for Non-Holonomic Agents\\ with Value Iteration Networks
}

\author{Eike Rehder$^{1,2*}$, Maximilian Naumann$^{3*}$, Niels Ole Salscheider$^{3}$, and Christoph Stiller$^{2,3}$%
\thanks{*Joint contribution}%
\thanks{$^{1}$Environment Perception, Daimler R\&D, Sindelfingen, Germany 
        {\tt\small eike.rehder@daimler.com}}%
\thanks{$^{2}$Karlsruhe Institute of Technology (KIT),
		Institute of Measurement and Control Systems, 76131 Karlsruhe, Germany
        {\tt\small stiller@kit.edu}}%
\thanks{$^{3}$FZI Research Center for Information Technology,
        Mobile Perception Systems, 76131 Karlsruhe, Germany
        {\tt\small \{naumann,salscheider\}@fzi.de}}%
}

\begin{document}

\maketitle
\thispagestyle{empty}
\pagestyle{empty}

\begin{abstract}

Cooperative motion planning is still a challenging task for robots. 
Recently, Value Iteration Networks (VINs) were proposed to model motion planning tasks as Neural Networks. In this work, we extend VINs to solve cooperative planning tasks under non-holonomic constraints. For this, we interconnect multiple VINs to pay respect to each other's outputs. Policies for cooperation are generated via iterative gradient descend. Validation in simulation shows that the resulting networks can resolve non-holonomic motion planning problems that require cooperation.

\end{abstract}

\section{Introduction}
\label{introduction}

Coordination of multiple robots is required whenever they share their workspace, so it is inevitable for most applications of mobile robotics.  This becomes tremendously hard for robots with non-holonomic constraints, i.e. constraints that are not expressible as a function of generalized coordinates. These constraints are for example imposed by an Ackermann steering geometry.

In many tasks in mobile robotics, the state space is continuous. 
However, often it is advantageous to discretize that space into small portions. 
Each of them then represents a fraction of the continuous space but may be addressed as a discrete variable. 
One popular form is the grid map in which a planar space is divided into equally spaced grid cells \cite{Elfes1989grids}.
Interpreting cells as nodes in a graph, motion planning can efficiently be solved using value iteration or quadratic graph search algorithms.
The latter can be simple, such as \textit{breadth first} or \textit{depth first}, or more sophisticated, such as \textit{Dijkstra’s algorithm}~\cite{dijkstra1959note} or \textit{A*}.

An alternative to the discretization of a continuous state space is its random exploration via sampling.
A well-known approach is the \textit{RRT*}~\cite{Karaman2011rrt_star}.

For some applications, the complexity of the motion planning problem can be reduced by separating the path planning from the velocity planning.
This approach, called path-velocity decomposition (PVD)~\cite{kant1986PVD}, is specially suited for static environments. 
However, PVD becomes sub-optimal for dynamic scenarios such as road traffic~\cite{bender2015combinatorial}.

Coordinating multiple robots adds a second level of difficulty.
While the configuration space of a single robot may already be large, it grows linearly with the number of robots.
As complete algorithms' computation time is at least exponential in configuration space dimension~\cite{lavalle2006planning}%
, their use without adaptation might be inappropriate.
Consequently, many approaches reduce the configuration space, e.g. by applying PVD and considering fixed paths for the multi robot coordination \cite{peng2003coordinating,altche2016time}.

Obviously, solving such complex tasks is a trade-off between the deviation from optimality by simplifying the task, and computational effort.

Recently, path planning for a single holonomic robot was proposed by \textit{Imitation Learning} of human behavior using a CNN \cite{tamar2016VIN, rehder2017imitationLearning}. 
Only based on observed trajectories of cars in urban environments, a CNN generated paths from previously unseen aerial images. However, this work could only deal with a single holonomic vehicle in a static environment. Real traffic, however, demands both, interaction and non-holonomic constraints.

In this work, we propose cooperative trajectory planning for multiple agents via gradient descent optimization.
The trajectory planning task for a single (non-)holonomic robot is solved using a Value Iteration Network.
Gradient descent optimization is realized by imposing mutual cost to each Value Iteration Network.
As input, we only define the static occupancy grid, a collision punishment and the agents' initial states and goals.
We demonstrate the performance of the approach by solving cooperative motion planning tasks for automated vehicles.

\begin{figure}%
\includegraphics[trim={0 0.5cm 0 0.5cm},clip,width=\columnwidth]{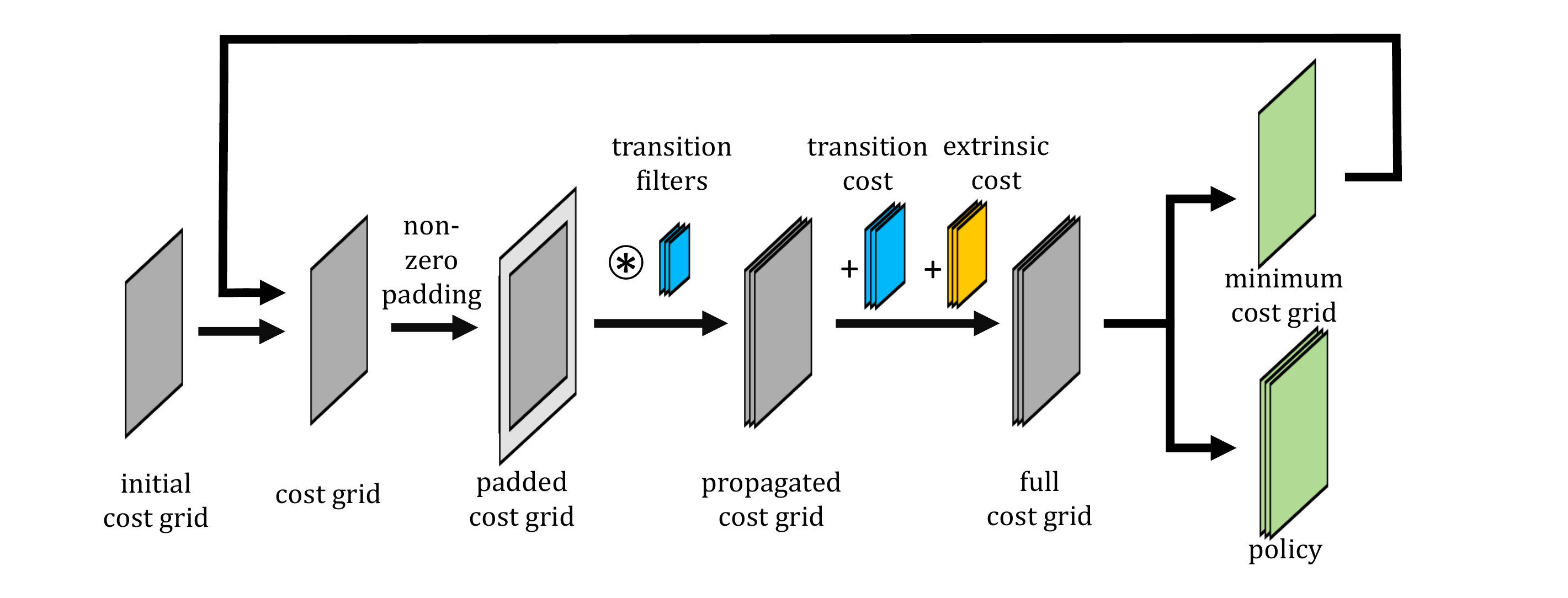} 
\caption{Value Iteration using a Convolutional Neural Network.}
\label{fig:VIN_cost_calc}
\end{figure}

\section{Background}\label{sec:background}
In this section, we provide background on discrete optimal planning in graphs and its implementation using CNNs.

\subsection{Discrete Optimal Planning}
This task of finding an optimal plan from a start node to a set of goal nodes in a graph is called \textit{Discrete Optimal Planning}.
This task can be solved using \textit{Value Iteration (VI)}.
Value Iteration iteratively computes \textit{cost-to-go} by propagating the cost from the previous state and adding the state transition cost.

In the following, we make use of the notation of LaValle~\cite{lavalle2006planning}:
\begin{itemize}
 \item $X$ denotes the \textit{state space}, containing all possible states
 \item $x \in X$ denotes a specific state
 \item $x_\mathrm{I}$ denotes the initial state
 \item $X_\mathrm{G} \subset X$ denotes the set of desired goal states.
\end{itemize}
Transitions in-between those states $x$ are possible through an action $u$ that is part of the \textit{action space} $u \in U$, containing all possible actions.
The transitions themselves from a state $x$ to a state $x'$ are defined by the \textit{state transition function} $f: X \rightarrow X$ with the \textit{state transition equation} $x' = f(x,u)$.

The value iteration for discrete optimal planning can be performed in different ways: 
Starting at the initial state $x_\mathrm{I}$, optimal \textit{cost-to-come} to any node can be calculated.
This approach is called \textit{forward value iteration}.
On the other hand, via \textit{backward value iteration}, optimal \textit{cost-to-go} to any goal state $x \in X_\mathrm{G}$ can be calculated.

Either approach yields the same (optimal) plan from an initial state to a set of goal states. 
However, we only consider forward value iteration in the following, as it simultaneously yields multiple plans starting at the (invariant) current pose of an agent.

To allow for optimal plans of unspecified lengths, a special \textit{termination action} $u_\mathrm{T}$ is commonly introduced.
Once $u_\mathrm{T}$ is applied, it is repeated forever: no more cost accumulates and the state remains unchanged.

\subsection{Value Iteration using a CNN}
In case the actions $u$ can be represented by a convolution in a state grid, CNNs can be made use of to implement value iteration algorithms \cite{tamar2016VIN, rehder2017imitationLearning}.

The underlying network architecture is shortly explained as in \cite{rehder2017imitationLearning} along Fig. \ref{fig:VIN_cost_calc}.

\begin{enumerate}
	\item \textbf{Initialization} - The convolution starts an initial cost grid, represented by a grid with only one zero in the initial state and high values otherwise.
	\item \textbf{Cost Propagation} - This cost grid is iteratively convolved with a set of filtermasks, representing one action $u$ each, resulting in one propagated cost grid per action.
	\item \textbf{Cost Accumulation} - The transitions cost and cost imposed by visiting a specific state are added to the respective cost grid
	\item \textbf{Optimal Cost and Policy Determination} - The minimum cost to visit a state and the respective policy are calculated by $\mathrm{min\mbox{-}pooling}$ and calculating a $\mathrm{one\mbox{-}hot}$ representation of the best action using the $\arg \min$ of cost grid.
	\item \textbf{Recursion} - Reapply steps, starting at step 2) with the minimum cost grid until convergence is reached.
\end{enumerate}

To also allow for optimal plans of unspecified lengths, the termination action $u_\mathrm{T}$ is replaced by imposing zero cost in the goal states.

The policy can now be evaluated as follows, along Fig. \ref{fig:NH_policy_execution}:

\begin{enumerate}
	\item \textbf{Initialization} - Initialize the state grid with a one in the desired goal state. %
	Reverse the transition filters by mirroring around the center point, as the reverse action has to be applied when backtracking.
	\item \textbf{Transition Selection} - Multiply the current state grid with the policy derived during the forward value iteration.
	The output is a $\mathrm{one\mbox{-}hot}$ representation of the action to be performed to reach this state.
	\item \textbf{State Propagation} - Reverse the detected action by convolving with the reversed transition filters.
	The output is the state from which the input state of step 2) can be reached with minimum cost.
	\item \textbf{Recursion} - Reapply steps, starting at step 2) with the output of step 4) until the initial state is reached.
\end{enumerate}

For local optimization, as well as for Imitation Learning (IL) and Reinforcement Learning (RL), full differentiability is beneficial.
In the upper algorithm, this can easily be achieved by replacing the $\arg \min$ in cost determination  by a $\mathrm{softmin}$ operation.
This modification still yields the optimal cost grid. 
However, the policy no longer represents the optimal one.
Instead, it only approximates the optimal policy but therefore provides a gradient.
For this reason, \textit{Value Iteration Networks} (VIN), as introduced by Tamar et al., are said to perform an approximate VI
computation \cite{tamar2016VIN}.

The advantages of CNN value iteration algorithms are twofold:

The exact implementation allows for use of well-established CNN frameworks. This facilitates fast discrete optimal planning.

The approximative implementation enables the combination of planning with ordinary learning techniques in neural networks, such as IL from driven paths on aerial images \cite{rehder2017imitationLearning}.

\section{Trajectory planning using CNNs}
\label{sec:tp_cnn}

In this section, we derive the network architecture for cooperative trajectory planning.
First, we extend the previously introduced path planning architecture to the time domain, enabling trajectory planning.
Subsequently, we enhance the latter to non-holonomic systems by extending the state space.
Finally, we propose learning of mutual cost via gradient descent optimization in order to enable cooperative planning.

\begin{figure*}%
\vspace*{3mm}
\includegraphics[width=2.\columnwidth]{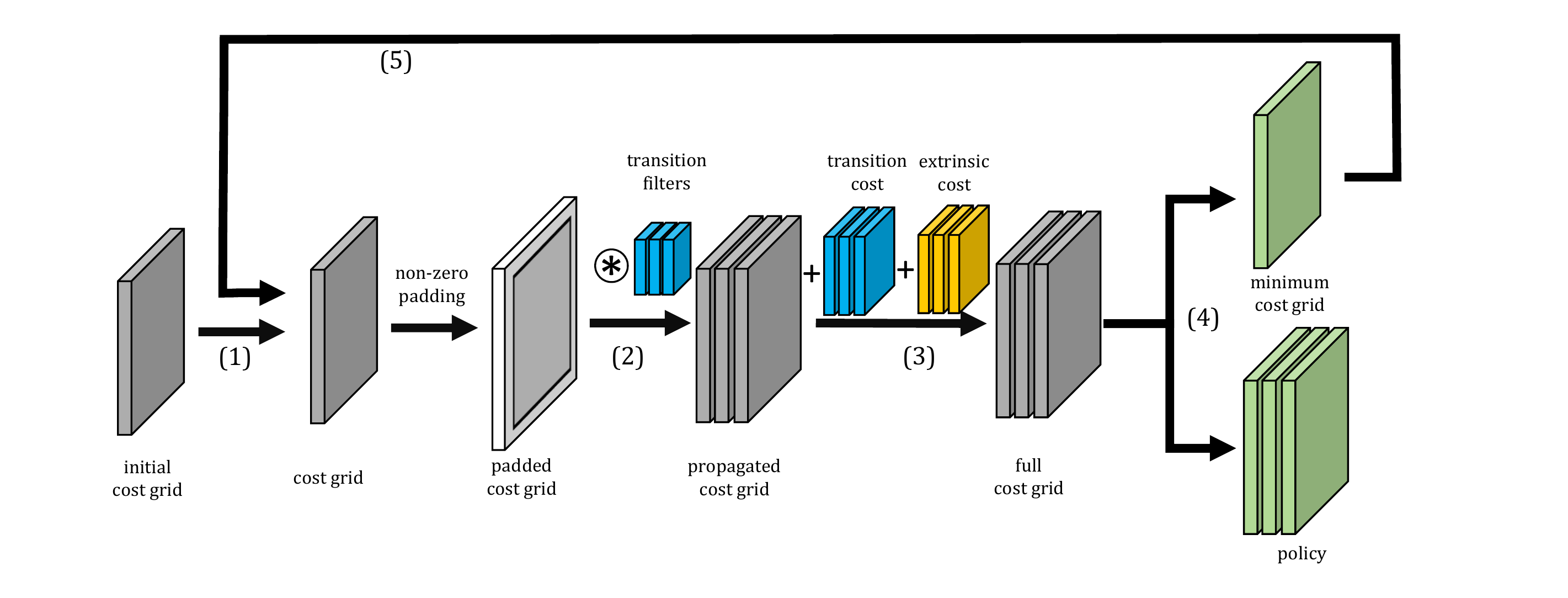} 
\caption{Policy Generation Network: Neural network to compute the cost grid and derive the motion policy. 
The convolution in (2) depicts a 2D-convolution of one state grid with a filter per possible transition and the concatenation of all transitions.
The step that is crucial to the behaviour is the summation of propagated cost, transition cost and extrinsic cost (3). 
The policy is derived from the result of the latter (4).}
\label{fig:NH_cost_grid_computation}
\end{figure*}

\subsection{Trajectory planning for holonomic robots using a CNN}

For trajectory planning, a time dimension has to be added to the path planning graph. 
As it is not possible to go back in time, the graph is directed in this dimension.
This can easily be achieved by assigning the iteration steps to time steps.
Thus, dynamic cost, for instance caused by dynamic obstacles, can be imposed.
The latter is substantial for problems involving multiple robots where collision free paths are suboptimal to collision-free trajectories on overlapping paths.

Referring to Fig. \ref{fig:VIN_cost_calc}, the initialization starts at $t=t_0$ and is incremented by one time step $\Delta t$ in cost determination until the planning horizon $t=t_\mathrm{max}$ is reached.
Consequently, the minimum cost grid is also time-dependent, representing both the reachability in time and the respective transition cost as well as dynamic cost.
Likewise, the policy is time-dependent.

\begin{figure*}%
\vspace*{3mm}
\includegraphics[width=2.\columnwidth]{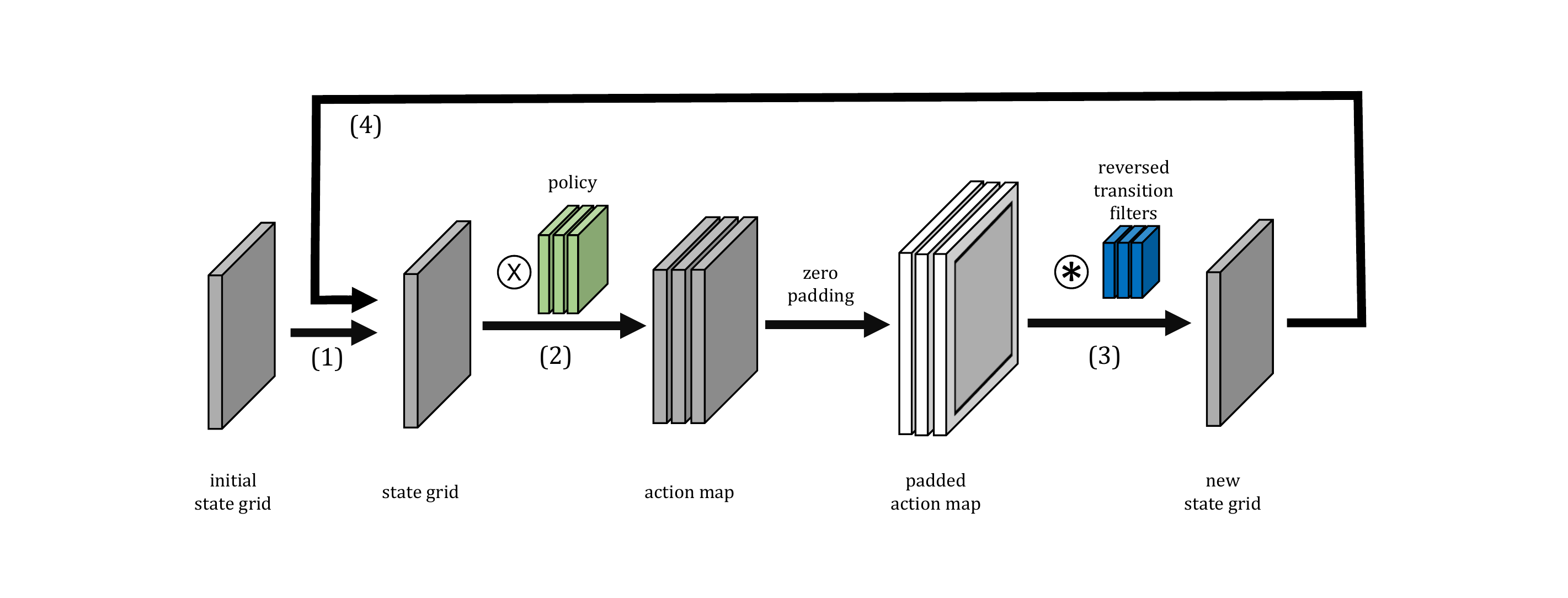} 
\caption{Policy Execution Network: Neural network to trace back the optimal trajectory. 
The policy is generated in the network depicted in Figure \ref{fig:NH_cost_grid_computation}. 
The multiplication in (2) denotes an element-wise multiplication of the state grid with one grid per transition and the concatenation of all transitions.}
\label{fig:NH_policy_execution}
\end{figure*}

For the evaluation of the policy, the state grid is initialized (cf. Fig. \ref{fig:NH_policy_execution}) in the desired state at the planning horizon $t=t_\mathrm{max}$.
Choosing the respective time-dependent policy, the state grid is backtraced to the initial state at time $t=t_0$.
For this initial state, no policy is available anymore, as the trajectory is complete.

As optimal plans of unspecified lengths are allowed, the cost added for the \textit{idle action} (stay in that state) in a goal state are zero.
This implies that, while backtracing from the goal state in $t=t_\mathrm{max}$, the idle action is the best action, as long as the goal state can be reached with similar cost already in an earlier timestep.
Only when reaching the goal state earlier becomes more expensive, the idle action is no longer chosen.
This is e.g. the case if the goal state is not reachable within a certain $t_\mathrm{n.r.}$.
Then, costs of this goal state are the standard infeasibility costs for all $t_0 \leq t \leq t_\mathrm{n.r.}$.
Consequently, if the goal state can be reached in $t < t_\mathrm{max}$, the trajectories that wait in the goal state instead of waiting in the initial state are automatically chosen by the policy.

In case of a dynamic obstacle along the path of a robot, all trajectories that proceed as soon as the obstacle vanishes have same cost.
Let's consider a 1d planning problem as example. Suppose we start at $x=0$ and end at $x=5$, and a dynamic obstacle is located in $x=4$ in timesteps $t_0 ... t_5$.
Theoretically, it is irrelevant, where in $x=0..3$ the robot waits until it reaches $x=4$ in $t_6$.
In practice however, we might want the robot to move towards the goal as soon as possible.
By doing so, we can reduce the risk of newly upcoming dynamic obstacles blocking our way.
Furthermore, if multiple agents share the same goal, they could overtake if there is still space in between the robot and the obstacle.
This behavior can be imposed by punishing the idle action more the earlier it is taken.
However, its cost should never overrule other action costs to prevent the robot from moving cyclic instead of waiting. 

Using an exact value iteration implementation, this approach yields optimal trajectories with respect to the defined cost.
Yet, using the approximative implementation in combination with RL or IL for learning the cost definition might outperform the upper approach with respect to "perceived" optimality, due to its superior, dynamic cost definition.

\subsection{Trajectory planning for non-holonomic robots using a CNN}

As some robots have kinematic constraints, e.g. car-like robots, the consideration of non-holonomic systems is required.
As the dynamics of non-holonomic motion cannot be reduced to generalized coordinates, additional state variables have to be introduced even for two-dimensional planning. 

In particular, in this work we consider the dynamics of a car-like robot. This class of robots features orientation-dependent state transitions since motion perpendicular to the wheel rotation is infeasible. Consequently, we incorporate the orientation into the state variables. In order to apply VINs %
to this model, we discretize the orientation in equidistant angular steps. 

While the original VIN operates on 2D state spaces, we now extend the transition function to three dimensions. The filter masks not only represent steps from one position to another. They rather describe orientation-dependent transitions. This way, we may exclude infeasible transitions such as sideway slip. This is in close relationship with the state lattice model for planning~\cite{ziegler2009lattices}. In the CNN case, the filter masks now represent the discrete set of transitions used in their planning works. 

From this point on, the dimensionality of all layers in the VIN increases by one: the cost grid becomes three dimensional. Since the transitions also map from 3D to 3D, the set of filter masks as well as the propagated grids now are 4D tensors. Finally, the min pooling operations to select the minimum cost per state reduces them to 3D again.

It is important to note in this context that the 4D tensors are fundamentally different from those used in standard CNNs. Usually, 4D tensors are used to represent a batch of 3D tensors that never exchange information. The fourth dimension only addresses the indexing of elements in the batch. In our work, the fourth dimension represents the action dimension and, thus, has to be handled accordingly. Figure \ref{fig:NH_cost_grid_computation} shows this switch compared to Figure \ref{fig:VIN_cost_calc}. Note how all layers now have changed to volumes where the width of a layer corresponds to the orientation dimension.

\subsection{Cooperative trajectory planning using gradient descent}

The challenge in cooperative trajectory planning via value iteration is that the graph grows exponentially with the number of agents.
To fully explore the graph, every action of every participant has to be combined iteratively.
In order to avoid this exhaustive computation, we impose time-dependent cost on each agent, caused by the designated trajectory of the respective other agent.
This time-dependent cost is determined via a gradient descent.
The objective function of the gradient descent consists of individual trajectory cost and collision cost, imposed by collisions of the agents.
The collision cost are defined as the sum of Frobenius inner products of orientation-independent state grids per timestep.
In order keep the agents from tunneling through one another, collisions with the previous state of the other agents are also punished.
Given that the combination of the initial trajectories of two agents lie in the same homotopy class, this gradient descent induces mutual cost in a way that the optimal trajectory ensemble is asymptotically obtained.  

In order to apply neural network training, full differentiability of the cost function is required. Thus, in training, we apply the softmin as explained in Section \ref{sec:background}. For execution of the trained policy, however, one would rely on hard decisions using min pooling.

\begin{figure}%
\includegraphics[width=0.95\linewidth]{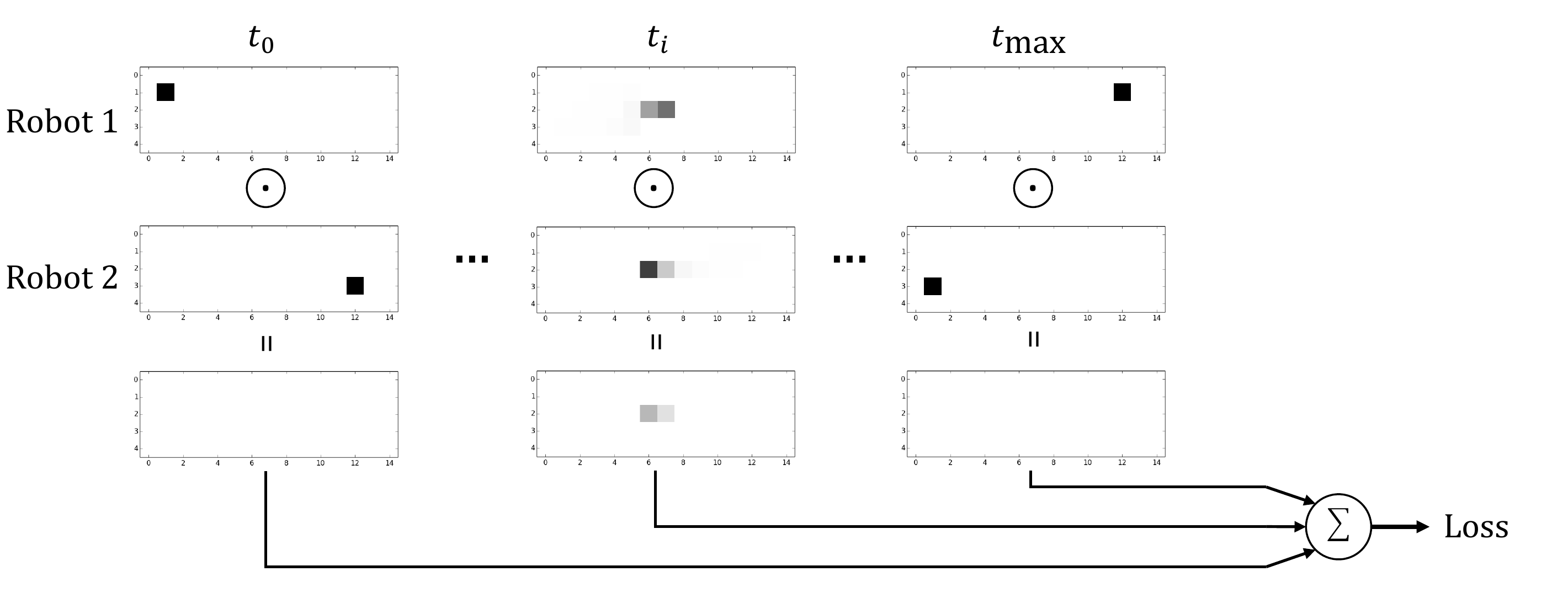} 
\caption{Loss function for cooperation: both policies are executed, then interconnected by time-step-wise Frobenius Inner Product}
\label{fig:cooperative_loss}
\end{figure}

\section{Experiments}
\label{sec:Experiments}

In the previous sections, we introduced non-holonomic cooperative planning using interconnected VINs. In this section, we show their performance in simulated robot motion experiments. Using a scenario with two robots facing a bottleneck, we demonstrate the applicability with different test cases. 

\begin{figure}  
\vspace*{3mm}
  \begin{subfigure}{\columnwidth}
  	\begin{center}
      \resizebox{0.92\linewidth}{!}{\input{case_5_car1.pgf}}
    \end{center}
    \vspace{-0.3cm}
    \caption{Trajectory of robot turning}
    \label{fig:case5}
  \end{subfigure}
  
  \vspace{0.8cm}
  \begin{subfigure}{\columnwidth}
   	\begin{center}
    \resizebox{0.92\linewidth}{!}{
\begingroup%
\makeatletter%
\begin{pgfpicture}%
\pgfpathrectangle{\pgfpointorigin}{\pgfqpoint{6.400000in}{1.147838in}}%
\pgfusepath{use as bounding box, clip}%
\begin{pgfscope}%
\pgfsetbuttcap%
\pgfsetmiterjoin%
\definecolor{currentfill}{rgb}{1.000000,1.000000,1.000000}%
\pgfsetfillcolor{currentfill}%
\pgfsetlinewidth{0.000000pt}%
\definecolor{currentstroke}{rgb}{1.000000,1.000000,1.000000}%
\pgfsetstrokecolor{currentstroke}%
\pgfsetdash{}{0pt}%
\pgfpathmoveto{\pgfqpoint{0.000000in}{0.000000in}}%
\pgfpathlineto{\pgfqpoint{6.400000in}{0.000000in}}%
\pgfpathlineto{\pgfqpoint{6.400000in}{1.147838in}}%
\pgfpathlineto{\pgfqpoint{0.000000in}{1.147838in}}%
\pgfpathclose%
\pgfusepath{fill}%
\end{pgfscope}%
\begin{pgfscope}%
\pgfsetbuttcap%
\pgfsetmiterjoin%
\definecolor{currentfill}{rgb}{1.000000,1.000000,1.000000}%
\pgfsetfillcolor{currentfill}%
\pgfsetlinewidth{0.000000pt}%
\definecolor{currentstroke}{rgb}{0.000000,0.000000,0.000000}%
\pgfsetstrokecolor{currentstroke}%
\pgfsetstrokeopacity{0.000000}%
\pgfsetdash{}{0pt}%
\pgfpathmoveto{\pgfqpoint{0.100000in}{0.703394in}}%
\pgfpathlineto{\pgfqpoint{6.300000in}{0.703394in}}%
\pgfpathlineto{\pgfqpoint{6.300000in}{1.047838in}}%
\pgfpathlineto{\pgfqpoint{0.100000in}{1.047838in}}%
\pgfpathclose%
\pgfusepath{fill}%
\end{pgfscope}%
\begin{pgfscope}%
\pgfpathrectangle{\pgfqpoint{0.100000in}{0.703394in}}{\pgfqpoint{6.200000in}{0.344444in}} %
\pgfusepath{clip}%
\pgftext[at=\pgfqpoint{0.100000in}{0.703394in},left,bottom]{\pgfimage[interpolate=true,width=6.210000in,height=0.360000in]{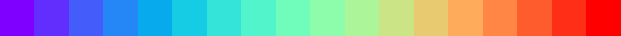}}%
\end{pgfscope}%
\begin{pgfscope}%
\pgfsetrectcap%
\pgfsetmiterjoin%
\pgfsetlinewidth{1.003750pt}%
\definecolor{currentstroke}{rgb}{0.000000,0.000000,0.000000}%
\pgfsetstrokecolor{currentstroke}%
\pgfsetdash{}{0pt}%
\pgfpathmoveto{\pgfqpoint{0.100000in}{1.047838in}}%
\pgfpathlineto{\pgfqpoint{6.300000in}{1.047838in}}%
\pgfusepath{stroke}%
\end{pgfscope}%
\begin{pgfscope}%
\pgfsetrectcap%
\pgfsetmiterjoin%
\pgfsetlinewidth{1.003750pt}%
\definecolor{currentstroke}{rgb}{0.000000,0.000000,0.000000}%
\pgfsetstrokecolor{currentstroke}%
\pgfsetdash{}{0pt}%
\pgfpathmoveto{\pgfqpoint{6.300000in}{0.703394in}}%
\pgfpathlineto{\pgfqpoint{6.300000in}{1.047838in}}%
\pgfusepath{stroke}%
\end{pgfscope}%
\begin{pgfscope}%
\pgfsetrectcap%
\pgfsetmiterjoin%
\pgfsetlinewidth{1.003750pt}%
\definecolor{currentstroke}{rgb}{0.000000,0.000000,0.000000}%
\pgfsetstrokecolor{currentstroke}%
\pgfsetdash{}{0pt}%
\pgfpathmoveto{\pgfqpoint{0.100000in}{0.703394in}}%
\pgfpathlineto{\pgfqpoint{6.300000in}{0.703394in}}%
\pgfusepath{stroke}%
\end{pgfscope}%
\begin{pgfscope}%
\pgfsetrectcap%
\pgfsetmiterjoin%
\pgfsetlinewidth{1.003750pt}%
\definecolor{currentstroke}{rgb}{0.000000,0.000000,0.000000}%
\pgfsetstrokecolor{currentstroke}%
\pgfsetdash{}{0pt}%
\pgfpathmoveto{\pgfqpoint{0.100000in}{0.703394in}}%
\pgfpathlineto{\pgfqpoint{0.100000in}{1.047838in}}%
\pgfusepath{stroke}%
\end{pgfscope}%
\begin{pgfscope}%
\pgfsetbuttcap%
\pgfsetroundjoin%
\definecolor{currentfill}{rgb}{0.000000,0.000000,0.000000}%
\pgfsetfillcolor{currentfill}%
\pgfsetlinewidth{0.501875pt}%
\definecolor{currentstroke}{rgb}{0.000000,0.000000,0.000000}%
\pgfsetstrokecolor{currentstroke}%
\pgfsetdash{}{0pt}%
\pgfsys@defobject{currentmarker}{\pgfqpoint{0.000000in}{0.000000in}}{\pgfqpoint{0.000000in}{0.055556in}}{%
\pgfpathmoveto{\pgfqpoint{0.000000in}{0.000000in}}%
\pgfpathlineto{\pgfqpoint{0.000000in}{0.055556in}}%
\pgfusepath{stroke,fill}%
}%
\begin{pgfscope}%
\pgfsys@transformshift{0.272222in}{0.703394in}%
\pgfsys@useobject{currentmarker}{}%
\end{pgfscope}%
\end{pgfscope}%
\begin{pgfscope}%
\pgfsetbuttcap%
\pgfsetroundjoin%
\definecolor{currentfill}{rgb}{0.000000,0.000000,0.000000}%
\pgfsetfillcolor{currentfill}%
\pgfsetlinewidth{0.501875pt}%
\definecolor{currentstroke}{rgb}{0.000000,0.000000,0.000000}%
\pgfsetstrokecolor{currentstroke}%
\pgfsetdash{}{0pt}%
\pgfsys@defobject{currentmarker}{\pgfqpoint{0.000000in}{-0.055556in}}{\pgfqpoint{0.000000in}{0.000000in}}{%
\pgfpathmoveto{\pgfqpoint{0.000000in}{0.000000in}}%
\pgfpathlineto{\pgfqpoint{0.000000in}{-0.055556in}}%
\pgfusepath{stroke,fill}%
}%
\begin{pgfscope}%
\pgfsys@transformshift{0.272222in}{1.047838in}%
\pgfsys@useobject{currentmarker}{}%
\end{pgfscope}%
\end{pgfscope}%
\begin{pgfscope}%
\pgftext[x=0.272222in,y=0.647838in,,top]{\sffamily\fontsize{20.000000}{24.000000}\selectfont \(\displaystyle 0\)}%
\end{pgfscope}%
\begin{pgfscope}%
\pgfsetbuttcap%
\pgfsetroundjoin%
\definecolor{currentfill}{rgb}{0.000000,0.000000,0.000000}%
\pgfsetfillcolor{currentfill}%
\pgfsetlinewidth{0.501875pt}%
\definecolor{currentstroke}{rgb}{0.000000,0.000000,0.000000}%
\pgfsetstrokecolor{currentstroke}%
\pgfsetdash{}{0pt}%
\pgfsys@defobject{currentmarker}{\pgfqpoint{0.000000in}{0.000000in}}{\pgfqpoint{0.000000in}{0.055556in}}{%
\pgfpathmoveto{\pgfqpoint{0.000000in}{0.000000in}}%
\pgfpathlineto{\pgfqpoint{0.000000in}{0.055556in}}%
\pgfusepath{stroke,fill}%
}%
\begin{pgfscope}%
\pgfsys@transformshift{1.994444in}{0.703394in}%
\pgfsys@useobject{currentmarker}{}%
\end{pgfscope}%
\end{pgfscope}%
\begin{pgfscope}%
\pgfsetbuttcap%
\pgfsetroundjoin%
\definecolor{currentfill}{rgb}{0.000000,0.000000,0.000000}%
\pgfsetfillcolor{currentfill}%
\pgfsetlinewidth{0.501875pt}%
\definecolor{currentstroke}{rgb}{0.000000,0.000000,0.000000}%
\pgfsetstrokecolor{currentstroke}%
\pgfsetdash{}{0pt}%
\pgfsys@defobject{currentmarker}{\pgfqpoint{0.000000in}{-0.055556in}}{\pgfqpoint{0.000000in}{0.000000in}}{%
\pgfpathmoveto{\pgfqpoint{0.000000in}{0.000000in}}%
\pgfpathlineto{\pgfqpoint{0.000000in}{-0.055556in}}%
\pgfusepath{stroke,fill}%
}%
\begin{pgfscope}%
\pgfsys@transformshift{1.994444in}{1.047838in}%
\pgfsys@useobject{currentmarker}{}%
\end{pgfscope}%
\end{pgfscope}%
\begin{pgfscope}%
\pgftext[x=1.994444in,y=0.647838in,,top]{\sffamily\fontsize{20.000000}{24.000000}\selectfont \(\displaystyle 5\)}%
\end{pgfscope}%
\begin{pgfscope}%
\pgfsetbuttcap%
\pgfsetroundjoin%
\definecolor{currentfill}{rgb}{0.000000,0.000000,0.000000}%
\pgfsetfillcolor{currentfill}%
\pgfsetlinewidth{0.501875pt}%
\definecolor{currentstroke}{rgb}{0.000000,0.000000,0.000000}%
\pgfsetstrokecolor{currentstroke}%
\pgfsetdash{}{0pt}%
\pgfsys@defobject{currentmarker}{\pgfqpoint{0.000000in}{0.000000in}}{\pgfqpoint{0.000000in}{0.055556in}}{%
\pgfpathmoveto{\pgfqpoint{0.000000in}{0.000000in}}%
\pgfpathlineto{\pgfqpoint{0.000000in}{0.055556in}}%
\pgfusepath{stroke,fill}%
}%
\begin{pgfscope}%
\pgfsys@transformshift{3.716667in}{0.703394in}%
\pgfsys@useobject{currentmarker}{}%
\end{pgfscope}%
\end{pgfscope}%
\begin{pgfscope}%
\pgfsetbuttcap%
\pgfsetroundjoin%
\definecolor{currentfill}{rgb}{0.000000,0.000000,0.000000}%
\pgfsetfillcolor{currentfill}%
\pgfsetlinewidth{0.501875pt}%
\definecolor{currentstroke}{rgb}{0.000000,0.000000,0.000000}%
\pgfsetstrokecolor{currentstroke}%
\pgfsetdash{}{0pt}%
\pgfsys@defobject{currentmarker}{\pgfqpoint{0.000000in}{-0.055556in}}{\pgfqpoint{0.000000in}{0.000000in}}{%
\pgfpathmoveto{\pgfqpoint{0.000000in}{0.000000in}}%
\pgfpathlineto{\pgfqpoint{0.000000in}{-0.055556in}}%
\pgfusepath{stroke,fill}%
}%
\begin{pgfscope}%
\pgfsys@transformshift{3.716667in}{1.047838in}%
\pgfsys@useobject{currentmarker}{}%
\end{pgfscope}%
\end{pgfscope}%
\begin{pgfscope}%
\pgftext[x=3.716667in,y=0.647838in,,top]{\sffamily\fontsize{20.000000}{24.000000}\selectfont \(\displaystyle 10\)}%
\end{pgfscope}%
\begin{pgfscope}%
\pgfsetbuttcap%
\pgfsetroundjoin%
\definecolor{currentfill}{rgb}{0.000000,0.000000,0.000000}%
\pgfsetfillcolor{currentfill}%
\pgfsetlinewidth{0.501875pt}%
\definecolor{currentstroke}{rgb}{0.000000,0.000000,0.000000}%
\pgfsetstrokecolor{currentstroke}%
\pgfsetdash{}{0pt}%
\pgfsys@defobject{currentmarker}{\pgfqpoint{0.000000in}{0.000000in}}{\pgfqpoint{0.000000in}{0.055556in}}{%
\pgfpathmoveto{\pgfqpoint{0.000000in}{0.000000in}}%
\pgfpathlineto{\pgfqpoint{0.000000in}{0.055556in}}%
\pgfusepath{stroke,fill}%
}%
\begin{pgfscope}%
\pgfsys@transformshift{5.438889in}{0.703394in}%
\pgfsys@useobject{currentmarker}{}%
\end{pgfscope}%
\end{pgfscope}%
\begin{pgfscope}%
\pgfsetbuttcap%
\pgfsetroundjoin%
\definecolor{currentfill}{rgb}{0.000000,0.000000,0.000000}%
\pgfsetfillcolor{currentfill}%
\pgfsetlinewidth{0.501875pt}%
\definecolor{currentstroke}{rgb}{0.000000,0.000000,0.000000}%
\pgfsetstrokecolor{currentstroke}%
\pgfsetdash{}{0pt}%
\pgfsys@defobject{currentmarker}{\pgfqpoint{0.000000in}{-0.055556in}}{\pgfqpoint{0.000000in}{0.000000in}}{%
\pgfpathmoveto{\pgfqpoint{0.000000in}{0.000000in}}%
\pgfpathlineto{\pgfqpoint{0.000000in}{-0.055556in}}%
\pgfusepath{stroke,fill}%
}%
\begin{pgfscope}%
\pgfsys@transformshift{5.438889in}{1.047838in}%
\pgfsys@useobject{currentmarker}{}%
\end{pgfscope}%
\end{pgfscope}%
\begin{pgfscope}%
\pgftext[x=5.438889in,y=0.647838in,,top]{\sffamily\fontsize{20.000000}{24.000000}\selectfont \(\displaystyle 15\)}%
\end{pgfscope}%
\begin{pgfscope}%
\pgftext[x=3.200000in,y=0.339197in,,top]{\sffamily\fontsize{20.000000}{24.000000}\selectfont \(\displaystyle t\)}%
\end{pgfscope}%
\end{pgfpicture}%
\makeatother%
\endgroup%
}
  \end{center}
  \vspace{-0.3cm}
  \caption{Colormap illustrating the time steps, see text Sec. \ref{sec:colormap}}
  \label{fig:colormap}
\end{subfigure}

    \caption{A non-holonomic agent performing a turnaround. Starting headed to the right with desired orientation to the left, it has to move circular. }
\end{figure}

\begin{figure}
	\begin{subfigure}{\columnwidth}
	  \begin{center}
	    \resizebox{0.92\linewidth}{!}{\input{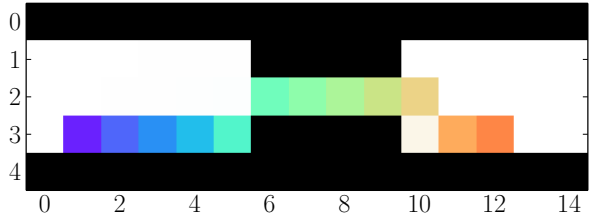}}
	  \end{center}
	  \vspace{-0.3cm}
	  \caption{Robot 1}
	\end{subfigure}
	
	\vspace{0.5cm}
	\begin{subfigure}{\columnwidth}
	  \begin{center}
	    \resizebox{0.92\linewidth}{!}{\input{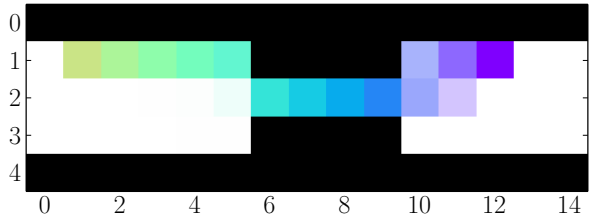}}
	  \end{center}
	  \vspace{-0.3cm}
	  \caption{Robot 2}
	\end{subfigure}
\caption{Two non-holonomic agents moving through a narrowing. The time is illustrated by a colormap (cf. Fig. \ref{fig:colormap}). The color denotes the latest time at which the agents are in a position.
Thus, robot 1 enters the narrowing only after robot 2 left it.}
\label{fig:case0}
\end{figure}

\begin{figure}
\vspace*{3mm}
	\begin{subfigure}{\columnwidth}
	  \begin{center}
	    \resizebox{0.92\linewidth}{!}{\input{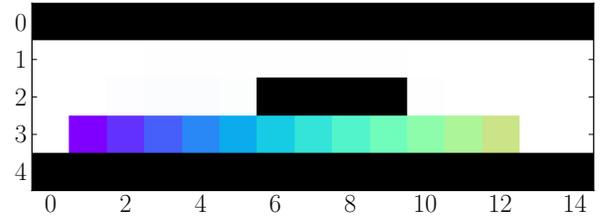}}
	  \end{center}
	  \vspace{-0.3cm}
	  \caption{Robot 1}
	\end{subfigure}
	
	\vspace{0.5cm}
	\begin{subfigure}{\columnwidth}
	  \begin{center}
	    \resizebox{0.92\linewidth}{!}{\input{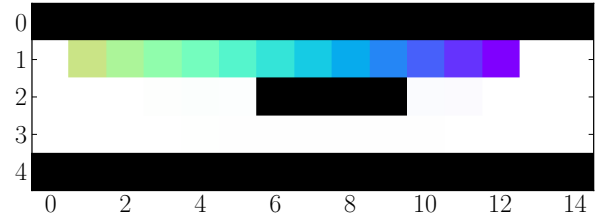}}
	  \end{center}
	  \vspace{-0.3cm}
	  \caption{Robot 2}
	\end{subfigure}
\caption{Planning without affecting each other if not necessary. The time is illustrated by a colormap (cf. Fig. \ref{fig:colormap}).}
\label{fig:case1}
\end{figure}

\begin{figure}
	\begin{subfigure}{\columnwidth}
	  \begin{center}
	    \resizebox{0.92\linewidth}{!}{\input{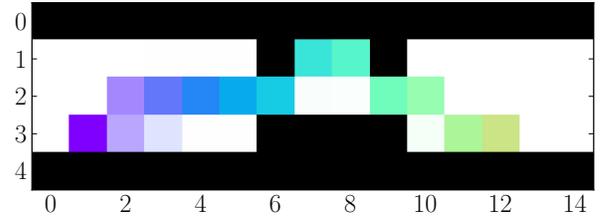}}
	  \end{center}
	  \vspace{-0.3cm}
	  \caption{Robot 1}
	\end{subfigure}
	
	\vspace{0.5cm}
	\begin{subfigure}{\columnwidth}
	  \begin{center}
	    \resizebox{0.92\linewidth}{!}{\input{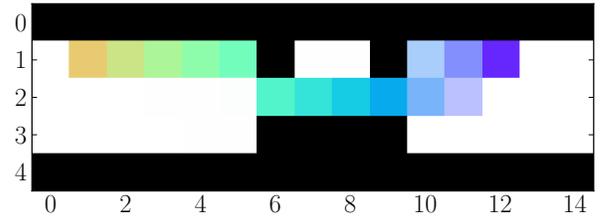}}
	  \end{center}
	  \vspace{-0.3cm}
	  \caption{Robot 2}
	\end{subfigure}
\caption{Navigating through a narrowing with a passing place. Robot 1 lets robot 2 pass in the narrowing. The time is illustrated by a colormap (cf. Fig. \ref{fig:colormap}).}
\label{fig:case2}
\end{figure}

\begin{figure}
\vspace*{3mm}
	\begin{subfigure}{\columnwidth}
	  \begin{center}
	    \resizebox{0.92\linewidth}{!}{\input{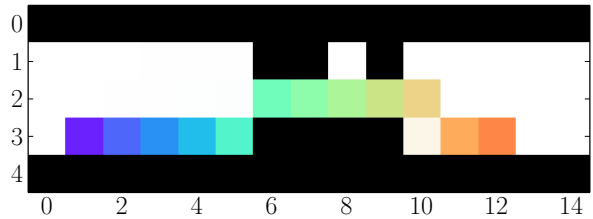}}
	  \end{center}
	  \vspace{-0.3cm}
	  \caption{Robot 1}
	\end{subfigure}
	
	\vspace{0.5cm}
	\begin{subfigure}{\columnwidth}
	  \begin{center}
	    \resizebox{0.92\linewidth}{!}{\input{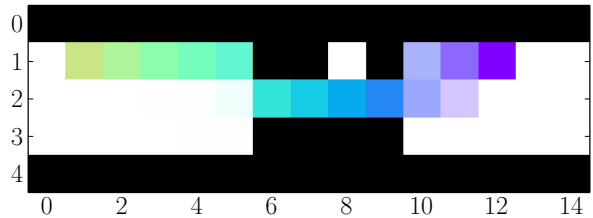}}
	  \end{center}
	  \vspace{-0.3cm}
	  \caption{Robot 2}
	\end{subfigure}
\caption{Navigating through a narrowing with a passing place that is not reachable for the robots due to their non-holonomic constraints. Thus, the result is similar to Fig. \ref{fig:case0}. The time is illustrated by a colormap (cf. Fig. \ref{fig:colormap}).}
\label{fig:case3}
\end{figure}

\subsection{Network Architecture}
The overall network architecture is described in-detail in Section \ref{sec:tp_cnn}.
Its implementation requires several design aspects: 
The size of the gridmap, the discretization of the additional dimension due to non-holonomicity, the choice of the transition filters and the respective costs as well as the choice of the extrinsic cost.

In our experiment we chose a gridmap of $15 \times 5$ cells and 8 orientations in $45^{\circ}$-steps each.
Further, we allow 4 actions per orientation, namely move straight, diagonal left, diagonal right and wait, represented by 32 transition filters. Note that diagonal steps also include a $45^{\circ}$-orientation change. 
The cost of the transitions are chosen to $1$ for vertical or horizontal movements in the grid and $\sqrt{2}$ for diagonal movements. 
Additionally, for every action, cost of $1$ is added to account for the time the action takes.
As explained earlier, no cost are imposed when waiting in the goal state.
The map is represented by a static cost grid, independent of the orientation.
Furthermore, the cost grid contains variable, orientation-independent cost that are initialized to zero and later adapted in the iterative gradient descent. 
The collision cost, calculated as explained in Section \ref{sec:tp_cnn}, are weighted with factor $100$ compared to the previously described transition cost. This high weighting serves as a soft constraint in planning.

\subsection{Visualization}\label{sec:colormap}
In the following, we display our results visually. Since we deal with trajectory planning in a discrete grid, we visualize trajectories in maps. Every map is a shown as a pixel image where each pixel represents one discrete cell. 

For every robot, we display the most relevant planning characteristics. We show the drivable area as white while static obstacles are black areas. The robot's trajectory is drawn as colored cells. Every color represents one single time step. The mapping from time to colors in shown in Fig.~\ref{fig:colormap}. If a robot occupies a single cell for multiple time steps, we only display the last. 

For cooperation scenarios, it is desired to avoid collision. In our visualization, this is equivalent to that same grid cells for the two robots must not feature same colors. 

Lastly, since we use the softmin to generate motion policies, different options per state may exist. Thus, we draw the softmin weight per trajectory cell as alpha channel of the color overlay. 

\subsection{Experimental Results}
In a first experiment, a single non-holonmic agent performs a turnaround by moving circular (cf. Fig. \ref{fig:case5}).

Cooperation is demonstrated with two non-holonomic agents passing a narrowing in opposite direction (cf. Fig. \ref{fig:case0}).
The narrowing can thereby only be passed by one agent at a time. 
The gradient descent yields that robot 2 passes first, as it is closer to the narrowing.

The second experiment shows, that the robots do not affect each other if not necessary (cf. Fig. \ref{fig:case1}).

In experiment C, the narrowing is extended by a passing place (cf. Fig. \ref{fig:case2}).
Thereby, robot 1 lets robot 2 pass in the passing place, and the overall cost are minimized.
If the narrowing is not reachable for the agents due to non-holonomic constraints, the solution is similar to the first experiment (cf. Fig. \ref{fig:case3}).

\section{Conclusions}
\label{sec:conclusion}

In this work, we proposed a neural network architecture to generate cooperative trajectories for multiple non-holonomic agents. First, we extended standard Value Iteration Networks to model the non-holonomic constraints, similar to state lattice models. Secondly, we interconnected and trained multiple VINs to display cooperative behaviour. 

In simulative experiments, we showed the applicability of these models to multi-agent motion planning. The networks were able to identify situations in which cooperation was necessary and act upon it. Also, if the situation could be resolved without interference, the networks found the corresponding solution. 

In our experiments, we assumed access to both agent's control. However, this rarely the case. Conveniently, the use of VINs doesn't stop there. As shown in previous works, VINs are also suitable for prediction \cite{rehder2017pedestrian}. Thus, if one of the agents was self-operated, we could simply exchange the deterministic transition filters in planning for probabilistic ones. This way, our work could readily be used for reactive motion planning. 

\section*{Acknowledgements}
This work was supported by the German collaborative research center ``SPP 1835 - Cooperative Interacting Automobiles'' (CoInCar) granted by the German Research Foundation (DFG) and the Tech Center a-drive. 

\bibliographystyle{IEEEtran}
\bibliography{references.bib}

\end{document}